\documentclass[conference]{IEEEtran}
\IEEEoverridecommandlockouts
\usepackage{cite}
\usepackage{amsmath,amssymb,amsfonts}
\usepackage{graphicx}
\usepackage{textcomp}
\usepackage{xcolor}
\usepackage{amsthm}
\usepackage{wrapfig,lipsum}
\usepackage{fancyhdr}       
\usepackage{adjustbox}
\usepackage{caption}
\usepackage{color}
\usepackage{mathtools,mathptmx}
\usepackage{newtxtext,newtxmath}
\usepackage{hyperref}
\usepackage{import}
\usepackage{algpseudocode}
\usepackage{algorithm}
\usepackage{subcaption}
\usepackage{orcidlink}

\usepackage[inkscapelatex=false]{svg}
\usepackage{booktabs}
\newtheorem{theorem}{Theorem}

\makeatletter
\newcommand{\linebreakand}{%
  \end{@IEEEauthorhalign}
  \hfill\mbox{}\par
  \mbox{}\hfill\begin{@IEEEauthorhalign}
}
\makeatother

\def\BibTeX{{\rm B\kern-.05em{\sc i\kern-.025em b}\kern-.08em
    T\kern-.1667em\lower.7ex\hbox{E}\kern-.125emX}}
\begin{document}

\title{GEDI: A Graph-based End-to-end Data Imputation Framework
}


\author{\IEEEauthorblockN{1\textsuperscript{st} Katrina Chen \orcidlink{0000-0001-9962-2974}}
\IEEEauthorblockA{
\textit{Dept. of
Statistical And Actuarial Science} \\
\textit{University Of Waterloo}\\
Waterloo, Canada \\
j385chen@uwaterloo.ca \\
}
\and
\IEEEauthorblockN{2\textsuperscript{nd} Xiuqin Liang}
\IEEEauthorblockA{
\textit{Data Science Center Of Excellence} \\
\textit{Deloitte Consulting}\\
Beijing, China \\
pliang@deloitte.com.cn} 
\and
\IEEEauthorblockN{3\textsuperscript{rd} Zheng Ma \orcidlink{0009-0005-6449-5983}}
\IEEEauthorblockA{
\textit{Dept. of Computer Science} \\
\textit{University Of Waterloo}\\
Waterloo, Canada \\
z43ma@uwaterloo.ca \\
} 
\linebreakand
\IEEEauthorblockN{4\textsuperscript{th} Zhibin Zhang}
\IEEEauthorblockA{
\textit{Institute of Computing Technology} \\
\textit{Chinese Academy of Sciences}\\
Beijing, China \\
zhangzhibin@ict.ac.cn}
}

\maketitle

\begin{abstract}
Data imputation is an effective way to handle missing data, which is common in practical applications. In this study, we propose and test a novel data imputation process that achieves two important goals: (1) preserve the row-wise similarities among observations and column-wise contextual relationships among features in the feature matrix, and (2) tailor the imputation process to some specific downstream label prediction task. The proposed imputation process uses Transformer and graph structure learning to iteratively refine the contextual relationships among features and similarities among observations. Moreover, it implements a meta-learning framework to select features that are influential to the downstream prediction task of interest. We conduct experiments on real-world datasets, and show that the proposed method consistently improves imputation and label prediction performance over a variety of benchmark methods.
\end{abstract}


\section{Introduction}
Tabular data is a ubiquitous data type in many practical applications, including healthcare, e-commerce, finance, law, and bioinformatics~\cite{borisov2021deep}. However, it is common for tabular data to contain missing values. For example, customer profiles on an online advertising platform may lack attributes such as gender, occupation, or age, and medical test results may have missing values due to missed appointments or incomplete tests.

To handle missing data in tabular data, data imputation is a standard technique. This approach replaces missing values in the data based on observed data. The completed feature matrix is then used in downstream label prediction tasks such as classification or regression. We argue that a well-designed imputation model should consider two critical aspects: (1) capturing underlying relationships within the observed data, and (2) aligning with the goal of the downstream task. 

\subsection{Learning the Relationship among Features and among Observations}
\label{section: learning interaction}
Missing values in tabular data can be inferred through contextually similar features as well as other similar observations. For example, a user's shopping preferences can be inferred from their age, gender, or occupation, and from others who have similar profiles. Thus, both the column-wise relationship among features and the row-wise relationship among observations are crucial for improving imputation performance.

Existing data imputation approaches suffer from limitations such as restrictive capabilities to model the relationship among heterogeneous features (i.e., a mixture of continuous and categorical features) or a lack of explicit utilization of information from observations. Some statistical methods, such as EM~\cite{ghahramani1993supervised} and MICE~\cite{white2011multiple}, impose probabilistic assumptions that are too restrictive for mixed-modal data, while others, such as kNN~\cite{maillo2017knn} and matrix completion~\cite{cai2010singular}, cannot handle data with categorical or numerical types. Conversely, the majority of machine learning and deep learning methods, such as missForest~\cite{stekhoven2012missforest}, SVM~\cite{wang2006missing}, generative adversarial networks~\cite{yoon2018gain,mattei2019miwae,nazabal2020handling,ma2020vaem}, and denoising auto-encoders~\cite{gondara2018mida,vincent2008extracting,tihon2021daema,nazabal2020handling}, do not explicitly model the row-wise relationship among observations. These limitations highlight the need for improved imputation methods that can effectively capture both the column-wise and row-wise relationships in tabular data.

Recent studies have utilized the potential of graph neural networks (GNNs)~\cite{hamilton2017representation} to capture the row-wise and column-wise relationships within the data. Nevertheless, some of these approaches have limited scalability when handling large-scale datasets due to the inefficient graph structure employed (e.g.,~\cite{you2020handling}). Additionally, other methods rely on pre-existing or sub-optimal graph structures, which may not be suited for the specific target task~\cite{cini2021filling}~\cite{spinelli2020missing}. Furthermore, certain GNN-based models are not adaptable to tabular data containing mixed types of features~\cite{cini2021filling}.

\subsection{End-to-end Solution to Improve the Performance of Downstream Task}
\label{section:end-to-end introduction}
In addition to the limitations of existing imputation model architectures, the majority of approaches employ a two-step impute-then-predict strategy, which separates the imputation task from the downstream task. This approach can be computationally inefficient and suboptimal for the target task.

\begin{figure}[t!]
  \begin{center}
    \includegraphics[width=1\linewidth]{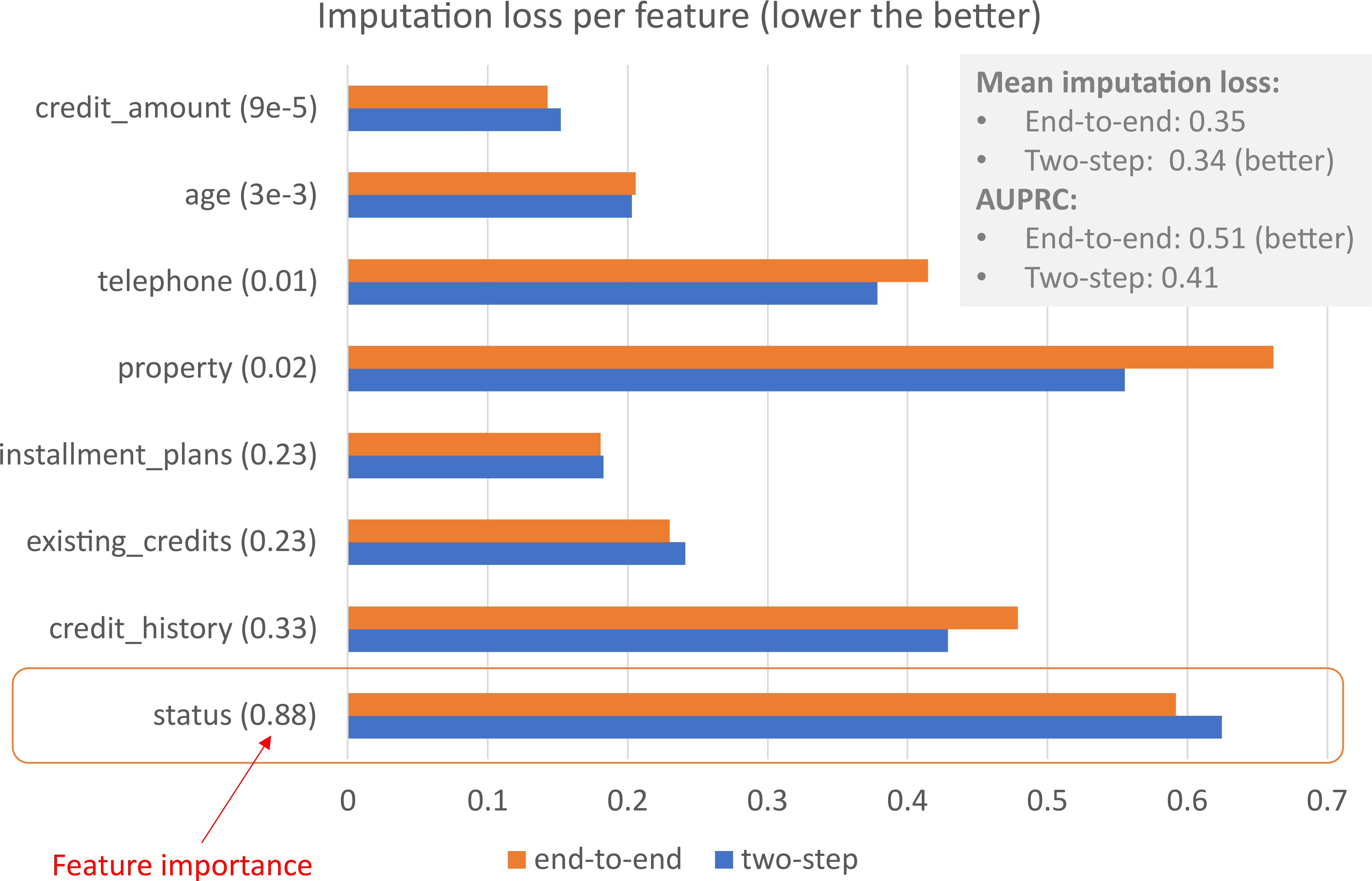}
  \end{center}
  \caption{Imputation loss of the top/bottom 4 features with the highest/lowest importance.}
   \label{fig:target_impute_consistency}
\end{figure}

Achieving higher imputation accuracy for features that are critical to the target task is preferable to simply minimizing the mean imputation loss, as features contribute differently to the task. This notion is illustrated through a simple experiment where we apply both the two-step impute-then-predict procedure and the proposed end-to-end framework to impute a dataset (credit-g from openml) with half of the entries missing at random. We subsequently apply logistic regression to the two imputed datasets to obtain both the AUPRC and feature importance, which reflects the absolute weights of each feature. As shown in Figure~\ref{fig:target_impute_consistency}, although the mean imputation error for the end-to-end method is higher than the two-step procedure, the former produces a better AUPRC. This result indicates that the end-to-end approach performs better in imputing features with higher importance. For instance, while the two-step procedure imputes better for the less important feature ``property'', it fares worse on ``status'', which has the highest importance.

\paragraph{Proposed}To address the limitations of existing approaches discussed in Sections~\ref{section: learning interaction} and~\ref{section:end-to-end introduction}, we introduce GEDI, a scalable end-to-end imputation framework designed to incorporate appropriate inductive biases into the model architecture while aligning with downstream tasks.

The main contributions of our work are as follows:

\begin{itemize}
\item We propose a deep learning architecture for the data imputation task that captures contextual relationships among features and similarities among observations efficiently, by leveraging graph structure learning and Transformer networks.
\item We apply a meta-learning framework to guide the model to focus on imputing the most influential features for the target task.
\item We conduct experiments on real-world datasets with continuous and categorical columns, and our results show that our model outperforms baseline methods in both data imputation and downstream target tasks.
\end{itemize}

\section{Literature Review}
Traditional statistical data imputation methods cover a wide range of methods such as kNN~\cite{maillo2017knn}, MICE~\cite{sterne2009multiple}, missForest~\cite{stekhoven2012missforest}, EM~\cite{dempster1977maximum}~\cite{ghahramani1993supervised}, matrix completion~\cite{cai2010singular}~\cite{mazumder2010spectral}~\cite{berg2017graph}~\cite{zheng2018spectral}, etc. ~\cite{little2019statistical} provides a comprehensive review of those methods. Statistical methods generally have solid mathematical foundations and good interpretability but have limited capability when working with the real-world large tabular dataset with mixed attributes. Deep learning methods are more flexible than statistical methods due to their superiority at modeling complex non-linear patterns and the ability for efficient end-to-end training~\cite{goodfellow2016deep}. The majority of deep learning imputation approaches are built on top of the denoising auto-encoders~\cite{gondara2018mida}~\cite{vincent2008extracting}~\cite{tihon2021daema}, and generative adversarial networks~\cite{nazabal2020handling}~\cite{yoon2018gain}~\cite{mattei2019miwae}~\cite{ma2020vaem}. However, all generative and auto-encoder-based methods do not explicitly model the connection between observations which is critical to impute missing data, as they only use the features of a single observation as input to generate predictions.   

Graph neural networks (GNNs) are superior at capturing complex relationships for graph-structured data. Its applicability depends on whether the optimal graph structure can be found for a given dataset. GRAPE~\cite{you2020handling} constructs a bipartite graph with observations as nodes and observed feature values as edges and explicitly applies GNNs to learn the interaction between observations and features. However, the graph is constructed in the way that every observed entry (or edge) is connected to all observations and feature nodes in the data, which is difficult to scale for large real-world datasets. ~\cite{spinelli2020missing} constructs a kNN sparse graph based on the pairwise distance of raw features to capture the relationship among observations. The graph structure is sub-optimal for the target task as it is constructed in the preprocessing stage. Another line of work (e.g.,~\cite{cini2021filling}) focuses on time series imputation, which is not particularly designed for tabular data with heterogeneous features. In contrast, our work leverages the attention mechanism to learn the contextual information within the features with mixed types and extract the row-wise observation information by constructing a similarity graph. The graph is sparse, homogeneous, and task-specific in nature, which is scalable, computationally efficient, and optimal for the target task. 


To our knowledge, limited works adopt end-to-end training to supervise the data imputation process. To name a few, GRAPE~\cite{you2020handling} formulates the label prediction task as a node-level prediction task so that both feature imputation and label prediction can be trained in an end-to-end fashion. HIVAE~\cite{nazabal2020handling} treats the target label as an additional feature so that the target prediction task is converted into an imputation task. Additionally, other machine learning models, such as tree-based~\cite{loh2011classification}~\cite{chen2015xgboost}~\cite{ke2017lightgbm}, and deep neural networks~\cite{smieja2018processing}~\cite{somepalli2021saint} are able to make a prediction in the presence of the missing values. Compared with those methods, our model can perform both data imputation and label prediction tasks and can be used as a plug-in component for deep neural network models to further improve their label prediction performance.  


\section{Methodology}
\subsection{Problem Definition}
We adopt common notation conventions throughout this paper. Specifically, $X_{i\cdot}$ and $X_{\cdot j}$ denote the $i$-th row and $j$-{th} column, respectively, of an arbitrary 2D matrix $X$. The $(i,j)$-th entry of $X$ is denoted by $X_{ij}$. $Y_{i}$ denotes the $i$-th element of a vector $Y$. The $l_2$-norm of a vector is denoted by ${\| \cdot \|}_2$. We use $\oslash$ to represent element-wise division, $\odot$ for element-wise multiplication, and $\mathbin\Vert$ for vector concatenation.

Given a tabular dataset $(X, Y)$, $X \in R^{N \times k}$ represents the input feature matrix containing $N$ observations with $k$ features, and $Y \in R^N$ denotes the target labels corresponding to the $N$ samples in $X$. We consider two tasks: (1) Data imputation task: Given a binary missing mask $M \in \{0,1\}^{N \times k}$  where $M_{ij}=1$ indicates that $X_{ij}$ can be observed, the goal is to predict the non-observed entries $X_{ij}$ when $M_{ij}=0$. (2) Label prediction task: Given a train/test partition $V \in \{0, 1\}^{N}$, the goal is to predict $Y_i$ when $V_i=0$.  

\subsection{Imputation Model}
In this section, we present the fundamental architecture of our imputation model, which comprises two key components: (1) a heterogeneous feature encoder that aims to capture the inter-column relationships among the heterogeneous features, and (2) a graph encoder that aims to learn the intra-row relationships among observations. The imputations are generated by leveraging both the row-wise and column-wise representations extracted from these two modules.

\begin{figure*}[t!]
\begin{subfigure}[t]{.38\textwidth}
  \centering
    \includegraphics[width=0.9\linewidth]{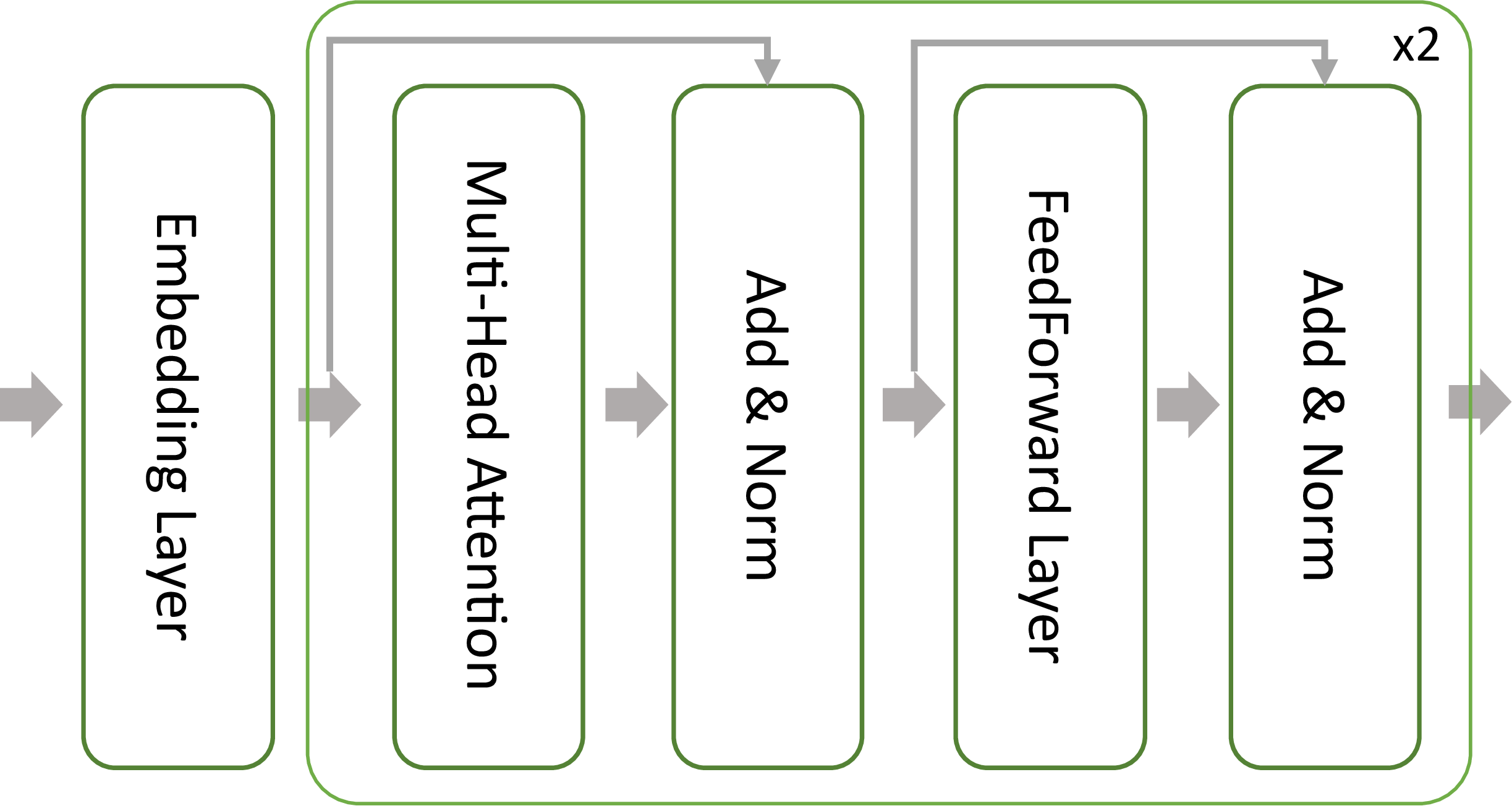}
  \caption{}
  \label{fig:feature_encoder}
\end{subfigure}
\hfill
\begin{subfigure}[t]{.58\textwidth}
    \includegraphics[width=0.9\linewidth]{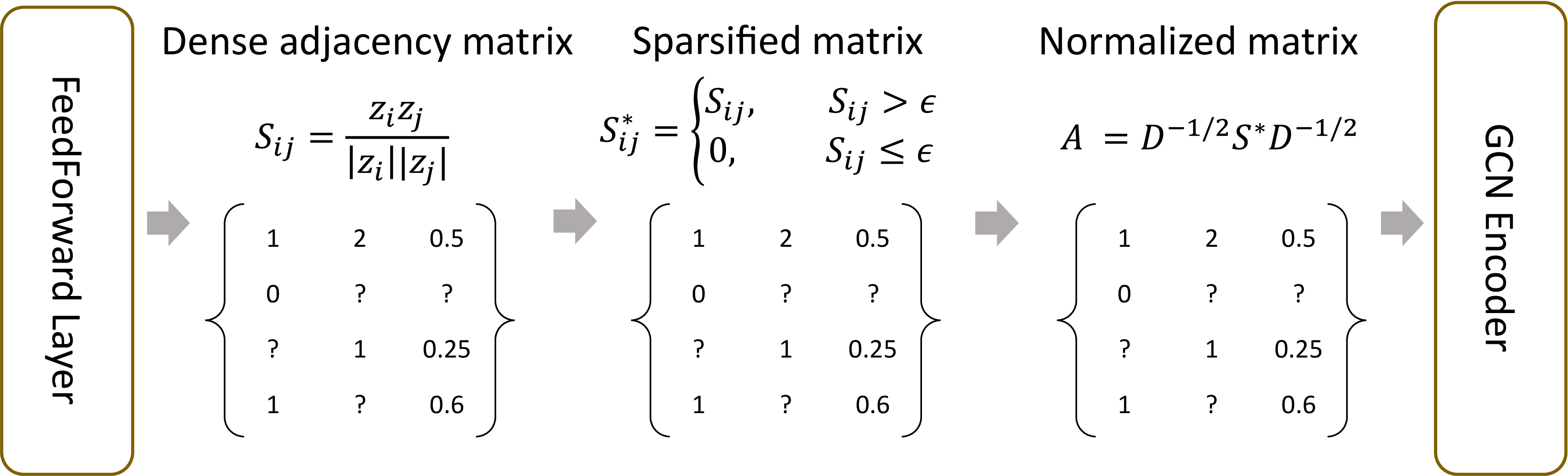}
  \caption{}
  \label{fig:graph_encoder}
\end{subfigure}
\newline
\begin{subfigure}[t]{1\textwidth}
  \centering
  \includegraphics[width=0.9\linewidth]{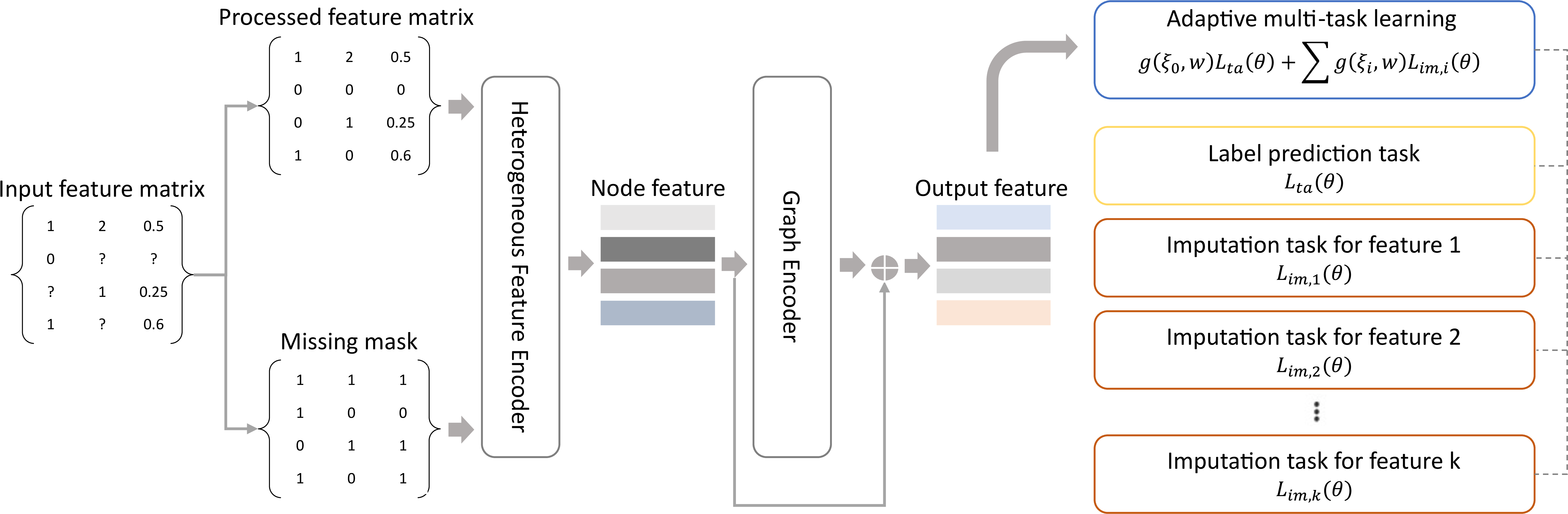}
  \caption{}
  \label{fig:imputation_model}
\end{subfigure}

\caption{Overall architecture and training pipeline of GEDI: (a) Feature encoder. (b) Graph encoder. (c) The complete end-to-end training process: models optimized under the multi-task training with weight update.}
\label{fig: framework}
\end{figure*}

\subsubsection{Heterogeneous Feature Encoder}
The feature encoder takes a feature matrix $X$ with missing entries filled with zero and its corresponding missing mask $M$ as input, and produces a dense feature matrix $Z$ in $R^{N \times d}$ as the output. Concretely, given $X_{i\cdot}$, the $i$-th row/observation from the input feature matrix, we first project each feature $X_{ij}$ into a $d$-dimensional feature space through an embedding layer. The classic embedding layer cannot handle continuous features directly. Therefore, we process each continuous feature separately by concatenating it with a one-hot vector (unique for each continuous feature) and then pass it through a one-layer feedforward neural network. The parameters of the layer are shared for all continuous features.

Next, we apply the Transformer model to learn the contextual information for each feature. We utilize the same encoder architecture as the original Transformer paper~\cite{vaswani2017attention}, without positional embeddings since there is no inherent ordering among the features. The Transformer block consists of a multi-head self-attention layer (MHA), a feedforward layer (FF), and residual connections~\cite{he2016deep}, followed by layer normalization (LN)~\cite{ba2016layer} (refer to Figure~\ref{fig:feature_encoder}). We feed all features into the Transformer with two stacked layers, and then apply a pooling layer to compute the mean representation of the outputs from the Transformer. Both the Transformer and pooling layers use the missing mask $M$ as the sequence attention mask and pooling mask respectively, ensuring that only $X_{ij}$ with $M_{ij}=1$ contribute to the final representation of the observation $X_{i\cdot}$. The full pipeline is
\begin{align}
    & R_{i\cdot} = \textnormal{LN}((\textnormal{MHA}(\textnormal{EB}(X_{i\cdot}), M_{i\cdot})+\textnormal{EB}(X_{i\cdot}))) \label{eq:first_transformer} \\
    & Z_{i\cdot} = \textnormal{Pooling}(\textnormal{LN}((\textnormal{MHA}(R_{i\cdot}, M_{i\cdot})+R_{i\cdot})), M_{i\cdot}) \label{eq:second_transformer} 
\end{align}

\subsubsection{Graph Encoder}
The graph encoder contains the graph generator and the GCN encoder (see Figure~\ref{fig:graph_encoder}). Given the output $Z \in R^{N \times d}$ from the feature encoder, The graph generator takes $Z$ as input and outputs an adjacency matrix $A \in R^{N \times N}$. The operations are defined as follows
\begin{align}
    & Z^{*} = \textnormal{ReLU}(ZW) \\
    & S_{ij} = \frac{Z^*_{i\cdot} {Z^*_{j\cdot}}^T}{\|Z^*_{i\cdot}\|_2 \|Z^*_{j\cdot}\|_2}\\
    & S^*_{ij} = \left \{
        \begin{aligned}
             & S_{ij}, \ S_{ij} > \epsilon \\
             & \ \ \ 0, \ S_{ij} \leq \epsilon
        \end{aligned}
        \right.  \\
    & A = D^{-1/2} S^* D^{-1/2} 
    \label{eq:graph generator operations}
\end{align}
where A is the normalized matrix based on the sparsified symmetric adjacency matrix $S^*$, which aims to prune noisy observations based on a threshold $\epsilon$, D is the degree matrix of $S^*$, and $W\in R^{d\times d}$ is a learnable weight matrix.

Next, the interaction between observations can be learned by graph neural networks through message passing and propagation based on the connection in A. We consider the simple GCN~\cite{kipf2016semi} layer as the base module in our implementation. However, any GNN layer which takes a node feature matrix and an adjacency matrix as input can be applied. Concretely, the graph encoder contains $L$ GCN layers, which takes $A$ from the graph generator and $Z$ from the feature encoder as inputs, and outputs a new feature matrix $G \in R^{N \times d}$ that can be used for both imputation and downstream tasks. The operation of each GCN layer is formulated as
\begin{align}
    & G^{(l+1)} = f(G^{(l)}, A) \label{eq:gnn_first_layer}\\
    & f(G^{(l)}, A) = \textnormal{ReLU}(AG^{(l)}W) \label{eq:gnn_second_layer}
\end{align}
where $A$ is the sparse weighted adjacency matrix computed by graph generator, $G^{(0)}$ denotes the output matrix $Z$ from the heterogeneous feature encoder, and $W\in R^{d\times d}$ is a trainable weight matrix.

Finally, given the feature matrix $G$ and $Z$ from graph encoder and feature encoder, the imputed feature matrix $\hat{X}$ is obtained as 
\begin{align}
    \hat{X} = \text{ReLU}((G \mathbin\Vert Z)W_1)W_2 
    \label{eq:merge_features}
\end{align}
where $W_1 \in R^{2d \times d}$ and $W_2 \in R^{d \times k'}$ are trainable weight matrices. Note that $k'\geq k$ as each predicted categorical feature is represented with the one-hot form (e.g., feature with $m$ categories is represented as a $m$-dimensional vector).

\subsubsection{Time complexity analysis}
In the previous sections, we assumed that the imputation model takes the entire feature matrix as input for the sake of annotation simplicity. However, in real-world applications with large datasets, a batch input matrix $X \in R^{B \times k}$ with batch size B can be sampled in each training iteration to improve training efficiency, where $B \ll N$. This reduces the time complexity of the graph encoder to $O(BBd+Bdd)$, where $d$ is the hidden dimension and $B \gg d$. Our experiments show that a smaller batch size compared to the dataset size can still produce satisfactory performance, as demonstrated in Figure~\ref{fig:imputation_performance_against_batchsize}. Moreover, the running time can be further reduced by employing the scalable graph structure learning implementation in~\cite{chen2020iterative}, which leverages an anchor approximation technique to reduce the time complexity from quadratic to linear.

\subsection{End-to-end Optimization Framework}

In section~\ref{section:end-to-end introduction}, we use an illustrative example to show that the two-step approach can perform worse than end-to-end training due to the lack of knowledge from the downstream tasks. The following theorem shows that the two-step impute-then-predict process leads to sub-optimal performance compared to the end-to-end training.
\begin{theorem}
Let $f_{\theta}:R^{N \times k} \rightarrow R^N$ and $g_{\omega}:R^{N \times k} \rightarrow R^{N \times k}$ be arbitrary neural network architectures with learnable parameters $\theta$ and $\omega$, $L_{\textnormal{ta}}: R^{N} \times R^N \rightarrow R$ and $L_{\textnormal{im}}: R^{N \times k} \times R^{N \times k} \rightarrow R$ be any arbitrary loss functions. Define $L_1$ as
\begin{align}
 & L_1 = \min_{\theta\in\Theta}{L}_{\textnormal{ta}}(f_{\theta}(g_{\omega^{*}}(X)), Y) \label{eq:two_step_imputation_loss}   
 \\
 & s.t. \ \omega^{*}=\arg\min_{\omega \in \Omega}{L}_{\textnormal{im}}(g_{\omega}(X'), X) \label{eq:two_step_imputation}
\end{align}
and $L_2$ as
\begin{equation}
L_2 = \min_{\omega,\theta} {L}_{\textnormal{ta}}(f_\theta(g_\omega(X)), Y)
\label{eq:end-to-end}
\end{equation}
where $X \in R^{N\times k}$ and $Y \in R^N$ are real matrices, and $X'$ is a copy of $X$ with arbitrary percent of values filled with zeros. Then the following statement holds:
\begin{align}
    L_1 \geq L_2
    \label{eq:l1<l2}
\end{align}
\end{theorem}
Proof:

Let $\theta^+ = \arg\min_{\theta\in\Theta}
{L}_{\textnormal{ta}}(f_{\theta}(g_{\omega^{*}}(X)), Y)$, then
\begin{align}
    L_1 & = {L}_{\textnormal{ta}}(f_{\theta^+}(g_{\omega^{*}}(X)), Y)  \\
    \label{Eq. objective of L1}
    & \geq min_\omega {L}_{\textnormal{ta}}(f_{\theta^+}(g_{\omega}(X)), Y) \\
    & \geq min_{\omega,\theta} {L}_{\textnormal{ta}}(f_{\theta}(g_{\omega}(X)), Y) \\
    & = L_2
    \label{Eq. bigger than L2}
\end{align}
which concludes the proof. \qed

By setting $X$ and $Y$ as the input feature matrix and label vector, $g_{\omega}$ and $f_{\theta}$ as the imputation model and downstream target predictor, and $L_{\text{ta}}$ and $L_{\text{im}}$ as the downstream target loss and feature reconstruction loss, the theorem implies that the end-to-end training scheme in Eq.~\eqref{eq:end-to-end} can achieve a lower training loss compared to the two-step approach in Eq.~\eqref{eq:two_step_imputation_loss}-\eqref{eq:two_step_imputation}.

However, we argue that imputation tasks can still be beneficial to the downstream tasks as the former can impose certain regularization on the latter. For instance, some contextual relationships among features captured by the imputation tasks can be helpful to the target tasks. Nevertheless, the joint optimization of both imputation tasks and downstream tasks can lead to the negative transfer problem, where the optimization of one task can hurt the other. To avoid this issue, we treat the weights of the imputation tasks as meta-parameters and adaptively optimize them based on the target task performance (the idea is motivated by the related works~\cite{shu2019meta}~\cite{ hwang2020self}).  On the one hand, essential features can be imputed better under the guidance of downstream tasks. On the other hand, the adaptive weighting mechanism prevents negative transfer from multi-task learning, where the performance of the primary task can be harmed by unhelpful auxiliary tasks.

Formally, the new joint loss with weight update is defined as
\begin{align}
    L_{\textnormal{joint}}(\theta) = g(\xi_0;w) L_{\textnormal{ta}}(\theta) + \sum_{i=1}^k g(\xi_i;w) L_{\textnormal{im}, i} (\theta)
    \label{eq:joint_loss_with_weight}
\end{align}
where $g$ is a one-layer feedforward neural network with trainable parameter $w$,  and $\xi_i$ is a concatenation of the one-hot embedding, label, and loss of the task type $i$ (i.e., $0$ is the index of the label prediction task, $1$ to $k$ are the indices for the imputation tasks). $L_{\textnormal{ta}}$ is the loss of the target task, and $L_{\textnormal{im},i}$ is the imputation loss of the feature $i$, which is the mean squared loss for continuous features and cross-entropy loss for categorical features. For a fair comparison with the two-step approaches that have logistic regression as the target predictor, we apply a one-layer neural network plus a sigmoid layer on top of the imputed feature matrix $\hat{X}$ to obtain the predicted target label $\hat{Y}$ for classification tasks. 

Two steps are conducted to optimize the model parameters iteratively. First, the weighting model parameter is updated with respect to the target loss
\begin{align}
    & \hat{\theta}^{l+1} = \theta^l - \alpha \nabla_\theta^l L_{\textnormal{joint}}(\theta^l; w^l)
        \label{eq:one_step_gradient_on_impute_model_hat} \\
    & w^{l+1} = w^l - \beta \nabla_w L_{\textnormal{ta}}(\hat{\theta}^{l+1}(w^l)) 
    \label{eq:one_step_gradient_on_weight_model}
\end{align}
where $w$ and $\theta$ denote the trainable parameters of the weight model and imputation model,  and $\alpha$ and $\beta$ are the learning rates for each of the models. Note that Eq.~\eqref{eq:one_step_gradient_on_weight_model} is equivalent to performing a one-step gradient update with respect to the target loss on the meta imputation model. 
In the second step, the parameter of the imputation model is updated with respect to the joint loss with $w$ fixed
\begin{align}
    \theta^{l+1} = \theta^l - \alpha \nabla_\theta^l L_{\textnormal{joint}}(\theta^l;  w^{l+1})
    \label{eq:one_step_gradient_on_impute_model}
\end{align}

Additionally, we adopt the cross-validation procedure in~\cite{hwang2020self} to stabilize the training: In each iteration, we split the label vector $Y$ in the training set by $C$ folds. Next, Eq.~\eqref{eq:one_step_gradient_on_weight_model} is performed based on the average target loss on each fold. To further speed up the training, we conduct Eq.~\eqref{eq:one_step_gradient_on_impute_model_hat}-~\eqref{eq:one_step_gradient_on_weight_model} for every five epochs and run Eq.~\eqref{eq:one_step_gradient_on_impute_model} with the weight parameter fixed within the five epochs. The resulting training time is almost equivalent to the multi-task training scheme but with a moderate improvement in the target performance.

\section{Experiment}



\subsection{Models}
We choose the following methods as benchmarks: 1. \textbf{Mean imputation}: The method imputes the categorical feature columns with the mode and continuous feature columns with the mean. 2. \textbf{$k$-nearest neighbors (kNN)}: We adopt the same implementation as in~\cite{jager2021benchmark} which applies scikit-learn's KNeighborsClassifier to impute the categorical features. 3. \textbf{MICE}~\cite{white2011multiple}: MICE treats each feature with missing values as a function of other features and fits a regressor on the created dataset. 4. \textbf{SVD}~\cite{troyanskaya2001missing}: A matrix completion-based method which iteratively imputes the missing values through low-rank SVD decomposition.  5. \textbf{GLFM}~\cite{valera2017automatic}: A Bayesian latent feature model designed for datasets with heterogeneous attributes. 6. \textbf{HIVAE}~\cite{nazabal2020handling}: A deep generative adversarial network that incorporates the likelihood of different variables to impute the heterogeneous features. 7. \textbf{GRAPE}~\cite{you2020handling}: A GNN-based framework that transforms the data imputation task into an edge prediction task and label prediction task into the node prediction task.

\subsection{Datasets}
We run the experiments on nine real-world tabular classification datasets from openML and Kaggle, which are adult, bank\_marketing(bank for brief), banknote, blastchar, breast, credit-g, defaultcredit, shoppers and spambase. Those datasets contain a mixture of categorical and numerical features. See Table~\ref{tab:datasets} for the summary statistics of the data. By following the setting in~\cite{nazabal2020handling}~\cite{you2020handling}, we introduce missingness to the data by multiplying the data feature matrix with a randomly generated binary missing mask. The experiment is conducted under the MCAR setting, where the missingness does not depend on the missing or complete values in the dataset. We leave the MAR and MNAR settings for future studies. 
\begin{table}[h!]
    \centering
    \setlength{\tabcolsep}{0.9mm}{
    \begin{tabular}{lcccccc}\\ \toprule
     Dataset     & \# Categorical & \# Continuous & Data size & \# Positive  & \# Negative \\ \midrule
     adult  &          8          &      6                & 30162         & 7508    &  22654       \\ 
     bank      & 10                  & 10            & 41188    & 4640     & 36548       \\ 
     banknote  &   0                   & 4         & 1372          &  610     & 762        \\ 
     blastchar  &  16           &   3       &  7043   & 1869      &  5174       \\ 
     breast  &  0                    &   30       & 569         & 212      &  357       \\ 
     credit-g   &   14                    &  6        & 1000           & 300      & 700         \\ 
     defaultcredit   &  3        &  20        &30000   &6636       & 23364        \\ 
     shoppers     &   7      &   10       & 12330         &  1908     & 10422   \\ 
     spambase   &   0                   & 57         & 4600          &  1812     &  2788        \\ \bottomrule
    \end{tabular}}
    \caption{Summary statistics for the datasets used in the experiments.}
    \label{tab:datasets}
\end{table}

\subsection{Model Configuration for GEDI}
The experiments for all datasets adopt the same parameter settings. For the imputation model, the Transformer of the feature encoder is fixed with 2 layers and 4 heads, the threshold $\epsilon$ is $0.8$ for the graph generator, the number of layers for GCN encoder is set to 1, and the hidden dimension in all components of the framework is fixed at $32$. For the weight update model, the embedding size of each task type is fixed at $8$. For the data imputation task, We train the model for $10,000$ epochs using the Adam optimizer with a learning rate at $0.001$. The training batch size is set to be $5,000$ for all datasets. For the target prediction task, We train the models for $5,000$ epochs using the Adam optimizer with a learning rate of $0.001$ for the imputation model and $0.005$ for the weight model. The batch size is fixed at $2,000$ for all target prediction experiments.  

\begin{table*}[t!]
    \setlength{\abovecaptionskip}{0cm}
    \centering
        \caption{Mean imputation error across all missing rates.} 

    \begin{tabular}{cccccccccc}\\ \toprule
    &       adult & bank\_marketing & banknote & blastchar & breast & credit-g & defaultcredit & shoppers & spambase \\ \midrule
   Mean     &       0.423 &          3.290 &    0.220 &     0.502 &  0.571 &    0.496 &         0.169 &    0.305 &    0.077 \\
   MICE     &       0.624 &          0.555 &    0.162 &     0.866 &  \textbf{0.069} &    0.776 &         0.182 &    0.470 &    0.070 \\
   SVD      &       0.681 &          1.941 &    0.220 &     0.906 &  0.184 &    0.805 &         0.211 &    0.494 &    0.081 \\
   kNN      &       0.299 &          0.312 &    0.198 &     0.441 &  0.161 &    0.356 &         0.133 &    0.283 &    0.067 \\
   GLFM     &       0.551 &          0.600 &    0.191 &     0.703 &  0.123 &    0.678 &         0.165 &    0.365 &    0.093 \\
   HIVAE    &       0.307 &            0.359 &    0.195 &     0.288 &  0.110 &    0.364 &           0.131 &    0.247 &    0.071 \\
   GRAPE    &       0.198 &          0.210 &    \textbf{0.135} &     0.220 &  0.092 &    0.353 &         0.117 &    0.229 &    0.068 \\ \midrule
   GEDI-G     &  0.243 & 0.241 & 0.169 & 0.248 & 0.133 & 0.338 & 0.115 &  0.239 & 0.069 \\   
   GEDI-F     &  0.195 & 0.201 & 0.143 & 0.210 & 0.075 & 0.337  & 0.126 &    0.216 & 0.066 \\  
   GEDI     &  \textbf{0.193} & \textbf{0.193} & 0.137 & \textbf{0.198} & 0.092 & \textbf{0.320} & \textbf{0.111} &    \textbf{0.214} & \textbf{0.063} \\
   
   \bottomrule
    \end{tabular}
    \label{tab:impute_result}
    \vspace{-0.3cm}
\end{table*}

\begin{table*}[t!]
    \setlength{\abovecaptionskip}{0cm}
    \centering
    \caption{Mean target performance (AUPRC) across all missing rates.}

    \begin{tabular}{cccccccccc}\\ \toprule
    &       adult & bank\_marketing & banknote & blastchar & breast & credit-g & defaultcredit & shoppers & spambase \\ \midrule
   Mean     &  0.419 &          0.364 &    0.900 &     0.598 &  0.749 &    0.454 &         0.293 &    0.417 &    0.924 \\
   MICE     &  0.438 &          0.513 &    0.922 &     0.643 &  0.968 &    0.497 &         0.333 &    0.425 &    0.919 \\
   SVD      &  0.403 &          0.444 &    0.873 &     0.611 &  0.857 &    0.504 &         0.302 &    0.421 &    0.900 \\
   kNN      &  0.437 &          0.490 &    0.883 &     0.628 &  0.901 &    0.496 &         0.325 &    0.412 &    0.915 \\
   GLFM     &  0.355 &          0.473 &    0.890 &     0.590 &  0.957 &    0.428 &         0.304 &    0.553 &    0.886 \\
   HIVAE    &  0.503 &          0.346 &    0.606 &     0.565 &  0.927 &     0.506 &           0.522 &    0.580 &    0.887 \\
   GRAPE    &  0.704 &          0.521 &    0.907 &     0.611 &  0.970 &    0.457 &         0.498 &    0.526 &    0.949 \\ \midrule
   two-step     &  0.481 & 0.521 & 0.951 & \textbf{0.657} & 0.965 & 0.506 & 0.343 &  0.434 & 0.934 \\   
   direct     &  0.739 & 0.543 & 0.928 & 0.615 & 0.921 & 0.490 & 0.506 &    0.571 & 0.951 \\ 
   multi-task     &  0.701 & 0.528 & 0.962 & 0.649 & 0.985 &  0.500 & 0.507 &    0.590 & 0.962 \\ 
    GEDI     &  \textbf{0.750} & \textbf{0.579} & \textbf{0.972} & 0.651 & \textbf{0.987} & \textbf{0.512} & \textbf{0.523} & \textbf{0.606} &    \textbf{0.964} \\ \bottomrule
    \end{tabular}
 
    \label{tab:target_result}
\end{table*}

\subsection{Data Imputation Task}


\subsubsection{Experiment setting}
\label{section:imputation_experiment}
GEDI is trained under the mean imputation loss under this setting to evaluate the performance of the imputation model. Given an input feature matrix $X \in R^{N \times k}$, a random test missing mask $M^t \in \{0, 1\}^{N \times k}$ 
 is created according to the missing rate of 0.1 to 0.5 (e.g., 0.1 represents 10 percent of values missing). Additionally, we generate a random validation missing mask $M^v \in \{0, 1\}^{N \times k}$ with a missing rate of 0.1 and the observed values in $X$ according to the missing mask $M^v \odot M^t$ is used to compute an imputation error for validation. All neural network models employ early stopping. 
 
The evaluation metric for the task is defined as the mean imputation error across all features, where the error of jth feature is calculated as normalized root mean square error for numerical features, accuracy error for categorical variables, and displacement error for ordinal variables (Following the work proposed in~\cite{nazabal2020handling}).

\subsubsection{Comparison with benchmark models} 
\begin{figure*}[t!]
  \centering
\begin{subfigure}[t]{.49\textwidth}
  \centering
\includegraphics[height=5.2cm]{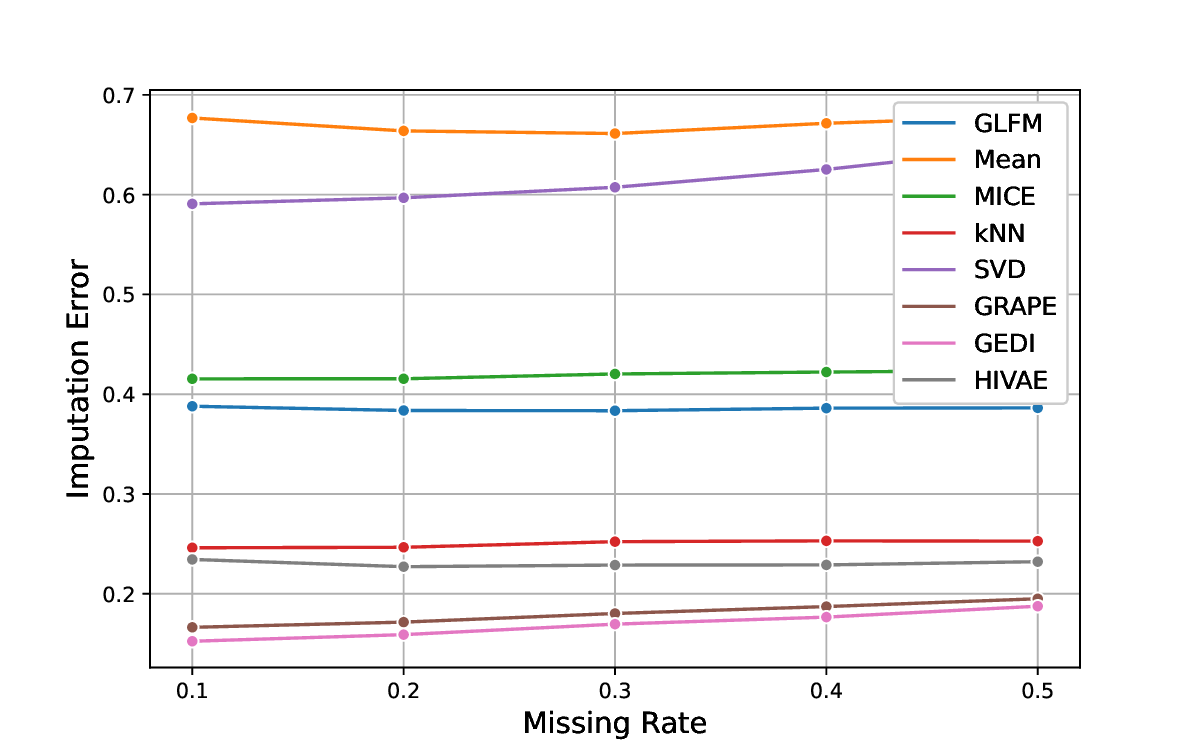}
\caption{}
\label{fig:imputation_performance_against_missing_rate}
\end{subfigure}
\hfill
\begin{subfigure}[t]{.49\textwidth}
  \centering
  {{\small}}
\includegraphics[height=5.2cm]{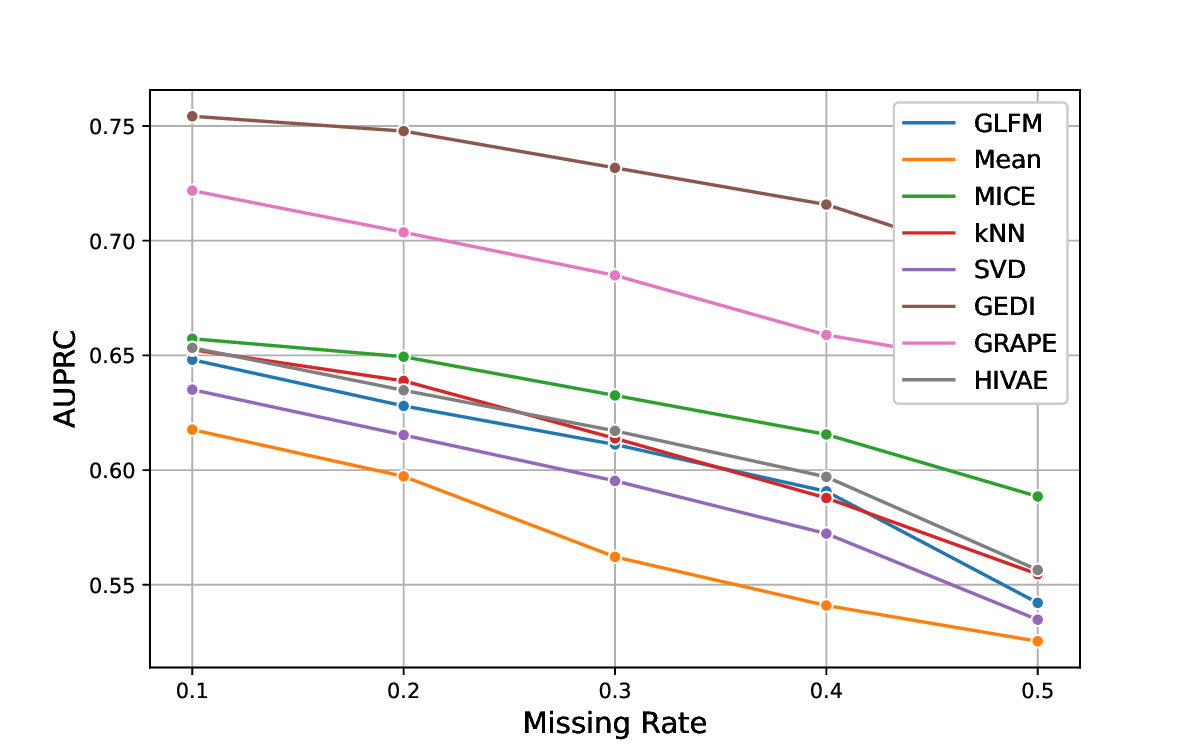}
\caption{}
\label{fig:target_performance_against_missing_rate}
\end{subfigure}
\newline
\begin{subfigure}[t]{.49\textwidth}
  \centering
\includegraphics[height=5.2cm]{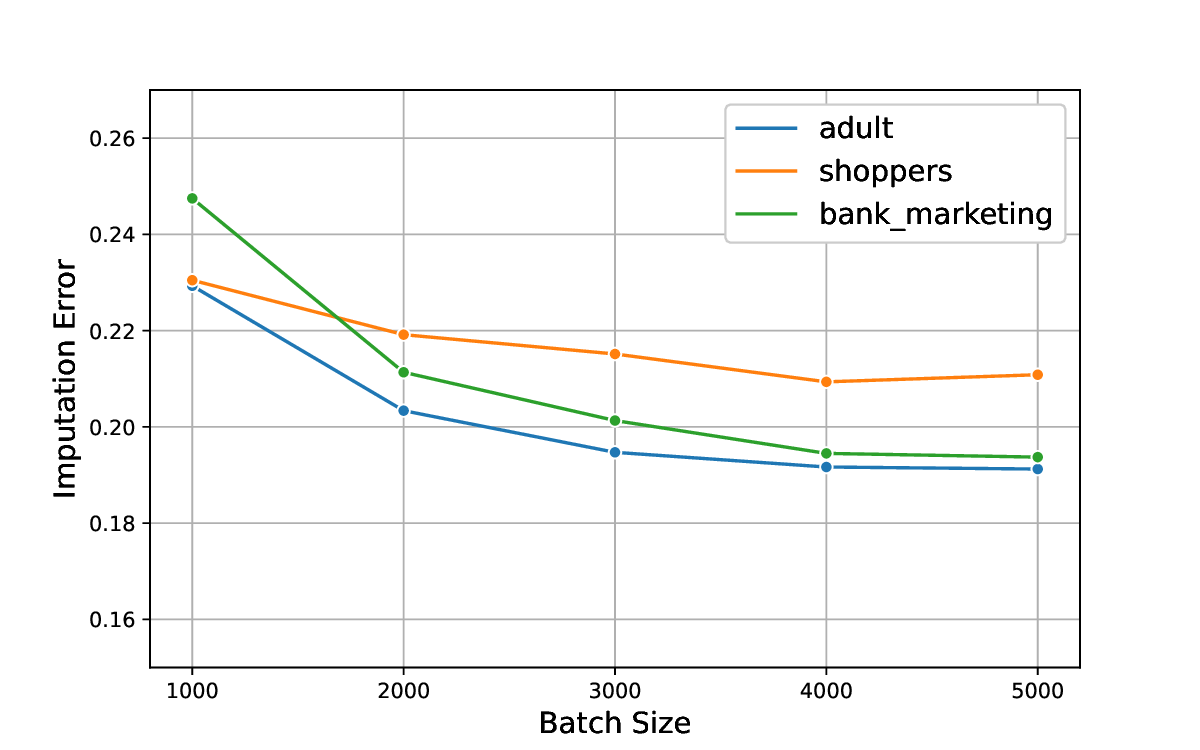}
\caption{}
\label{fig:imputation_performance_against_batchsize}
\end{subfigure}
\hfill
\begin{subfigure}[t]{.49\textwidth}
  \centering
\includegraphics[height=5.2cm]{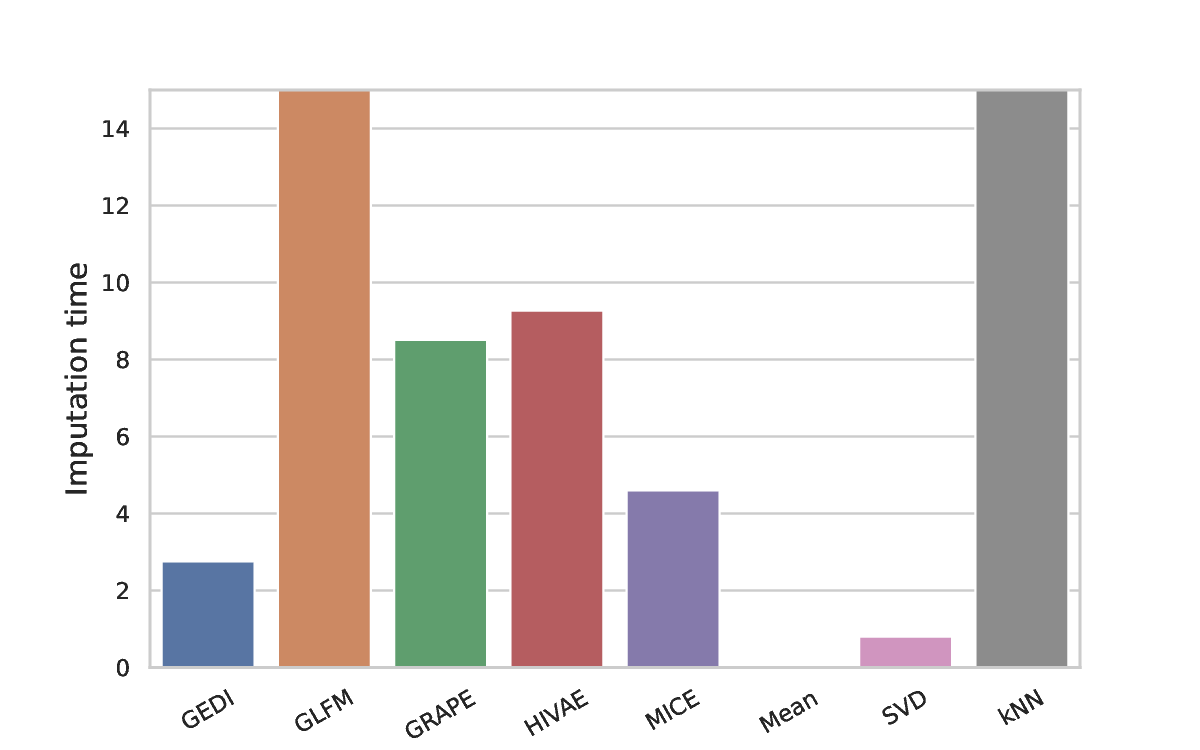}
\caption{}
\label{fig:imputation_time_against_methods}
\end{subfigure}
\caption{(a) Mean imputation error vs. missing rates.(b) AUPRC vs. missing rates. (c) Mean imputation error vs. batch size. (d) Mean imputation time in seconds among different methods.}
\label{fig: hyperparameter results}
\end{figure*}

Table~\ref{tab:impute_result} shows the mean imputation error for all datasets across missing rates, where GEDI has the lowest mean imputation error across the majority of the datasets. Note that GRAPE ranks higher than most methods, which further agrees with our analysis that the relationships among features and observations are critical for imputing the missing values since both GRAPE and GEDI incorporate this information explicitly into the model architecture. However, our model still outperforms GRAPE by 6$\%$ since GEDI has a more efficient graph structure to capture the similarity among observations. In addition, Figure~\ref{fig: hyperparameter results}a shows the robustness of models against different missing rates, where GEDI performs consistently across all missing rates. Apart from the competitive performance of graph-based imputation methods against other benchmarks, GEDI requires less than one-third of imputation time than GRAPE during the inference stage (Figure~\ref{fig: hyperparameter results}d). One reason is that GRAPE requires at least 2 to 3 layers of propagation of GNN to reach all the relevant features and observations (due to the interleaving bipartite architecture), while GEDI decouples the row-wise and column-wise representation learning into different components and only one layer of propagation is sufficient to aggregate the row-wise information based on the homogeneous similarity graph. 
Figure~\ref{fig: hyperparameter results}c also shows that the training batch size can be reduced to a certain point without sacrificing too much imputation performance, meaning that GEDI can be easily adapted to large datasets than GRAPE, which requires the complete datasets as input.

\subsubsection{Ablation study for imputation model}
We consider the following variants of the imputation model: GEDI-F has only the feature encoder, while GEDI-G has only the graph encoder with the node feature calculated based on the raw features. The bottom section of Table~\ref{tab:impute_result} shows the comparison of those variants, which verifies our assumption that both the dependencies among features and similarities among observations can help the model to make better predictions on the missing values.

\subsection{Label Prediction Task}

\subsubsection{Experiment setting}
GEDI is trained in the end-to-end mode, where the weight of the imputation losses is adaptively adjusted by the target task. Given a dataset with input feature matrix $X \in R^{N \times k}$ and a label vector $Y\in R^N$, we introduce the missing values to the feature matrix similarly to Section~\ref{section:imputation_experiment}. We create the train/test split according to the ratio $70/30\%$ and keep $20\%$ of the train set as the validation set. The parameter of all trainable models is stored at the epoch with the best performance on the validation set. The evaluation metric for the label prediction task is AUPRC since most of the datasets are highly imbalanced. For fair comparison, We apply logistic regression (one feedforward layer similar to GEDI) on top of the datasets imputed by models which do not support end-to-end training. Note that the aim of this study is not to compare with the classification models, but to demonstrate the performance of the imputation model as well as the end-to-end training scheme.

\subsubsection{Comparison with benchmark models}
Table~\ref{tab:target_result} shows the mean AUPRC performance against different methods. GEDI can achieve $8 \%$ improvement over the best-performing baseline, demonstrating the benefits of using the end-to-end framework. In comparison, the two-step impute-then-predict methods show the inconsistent performance between data imputation and target task (e.g., GEDI under the two-step approach achieves the lowest imputation error in 
shoppers but with an AUPRC lower than GLFM). Additionally, one observe that the end-to-end performance of HIVAE is even worse than the two-step imputation approaches. One reason is that HIVAE employs an indirect approach to convert the prediction task into a missing data imputation task, which does not include the data imputation as an intermediate step. In contrast, GRAPE and GEDI are built on top of the outputs of the imputation tasks and are further regularized by the imputation losses. The imputation process helps the prediction task learn the most valuable relationships among features and similarities among observations, which serves as a strong inductive bias to the prediction model to prevent overfitting. We also compare the label prediction performance against different missing rates in Figure \ref{fig: hyperparameter results}b, which shows the consistent and significant improvement of GEDI over the other baseline methods. 

\subsubsection{Ablation study for prediction framework}
We compare the performance for three variants of the proposed end-to-end framework: (1) two-step impute-then-predict process: the imputation and target task are trained separately; (2) direct training scheme: the imputation and target predictor models are optimized under the cross entropy loss of the target task. (3) multi-task training scheme: models are optimized under the cross entropy loss and the imputation loss with equal weights. The results are shown in the bottom section of Table~\ref{tab:target_result}. On average, the end-to-end training can significantly outperform the two-step impute-then-predict approach. However, the multi-task training scheme can suffer from the negative transfer problem, sometimes resulting in lower performance than direct training. The meta-weight framework boosts the performance upon multi-task training, verifying our assumption that emphasizing the imputation of the most influential subset of features can improve the performance of the downstream tasks.

\section{Conclusion}
Missing data is a prevalent issue for most real-world applications. In this work, we identify that the key to handling missing data is to capture the relationship among features and similarities among observations and align the goal of the missing data imputation task with the downstream task. We propose GEDI, a scalable data imputation framework that can iteratively incorporate the information flow among features and observations through graph structure learning to improve the performance of the missing data imputation task. Meanwhile, the framework is trained end-to-end so that the model can directly leverage the downstream information to improve the performance of the label prediction task, which is often the ultimate goal for real-world applications. Our empirical results show that GEDI can outperform the baseline imputation methods on real-world heterogeneous datasets in both data imputation and the downstream label prediction tasks.

\bibliographystyle{IEEEtran}
\bibliography{references}
\end{document}